\documentclass[journal]{IEEEtran}

\usepackage{subfigure}
\usepackage{amssymb}
\usepackage{graphicx}
\usepackage{algorithm}
\usepackage{algorithmic}
\usepackage{url}
\usepackage{multirow}
\usepackage{color}
\usepackage{amsthm}
\usepackage{amsmath}

\ifCLASSINFOpdf
\else
\fi
\begin{document}
%
\title{Active Dictionary Learning in Sparse Representation Based Classification}
%
%
%

\author{Jin~Xu,
        Haibo~He,~\IEEEmembership{Senior~Member,~IEEE,}
        and~Hong~Man,~\IEEEmembership{Senior~Member,~IEEE}
\thanks{J. Xu is with Operations and Information Management Department, University of Pennsylvania, Philadelphia, PA, 19104, USA e-mail: jinxu@wharton.upenn.edu}
\thanks{H. Man are with the Electrical and Computer Engineering Department, Stevens Institute of Technology,
Hoboken, New Jersey, 07030, USA e-mail: Hong.Man@stevens.edu.}
\thanks{H. He is with the Department of Electrical, Computer, and Biomedical Engineering, University of Rhode Island, Kingston, RI, 02881, USA e-mail: he@ele.uri.edu}

}

%
%

\markboth{PREPRINT SUBMITTED TO IEEE TRANSACTIONS ON NEURAL NETWORKS AND LEARNING SYSTEMS}%
{Shell \MakeLowercase{\textit{et al.}}: Bare Demo of IEEEtran.cls for Journals}
%



\maketitle

\begin{abstract}
Sparse representation, which uses dictionary atoms to reconstruct input vectors, has been studied intensively in recent years. A proper dictionary is a key for the success of sparse representation. In this paper, an active dictionary learning (ADL) method is introduced, in which classification error and reconstruction error are considered as the active learning criteria in selection of the atoms for dictionary construction. The learned dictionaries are caculated in sparse representation based classification (SRC). The classification accuracy and reconstruction error are used to evaluate the proposed dictionary learning method. The performance of the proposed dictionary learning method is compared with other methods, including unsupervised dictionary learning and whole-training-data dictionary. The experimental results based on the UCI data sets and face data set demonstrate the efficiency of the proposed method.
\end{abstract}

\begin{IEEEkeywords}
sparse representation; active learning; dictionary learning; classification; boosting; machine learning
\end{IEEEkeywords}

%
\IEEEpeerreviewmaketitle

\section{Introduction}
Sparse representation (sparse coding) \cite{proceeding0}, which uses a few atoms from a dictionary to construct a signal, is a popular method to acquire, compress and represent signals. In particular, a vector $y \in \mathbb{R}^{m}$ has a sparse representation $y=Dx$ on a dictionary $D \in \mathbb{R}^{m \times n}$, when the correspondent vector $x \in \mathbb{R}^{n}$ is sparse (the majority of the coefficients are zeros). There exist many successful applications using sparse representations, such as face recognition \cite{proceeding1}, image denoising \cite{jour1}, blind source separation \cite{jour11} and feature selection \cite{jinadd4}, \cite{jinadd5}.

It is well known that sparse representation is computational intensive \cite{jinadd3}. The property of the dictionary can affect the sparse representation significantly. How to construct a proper dictionary is important for sparse representation. There are two major approaches for dictionary learning. First is the analytic approach, in which DCT bases, wavelets, curvelets and other nonadaptive functions are used as atoms to construct the dictionaries. Second is the learning-based approaches, such as the unsupervised learning for dictionary construction \cite{proceeding5} and the online dictionary learning \cite{proceeding2}, \cite{jour51}, which use machine learning methods to construct the dictionary.

Active learning \cite{proceeding50}, \cite{proceeding51} is a machine learning \cite{jinadd1}, \cite{jinadd2} paradigm, in which the learning algorithm \cite{jinadd10} select certain unlabeled data for labeling through interactive queries with the user (or information source). It is generally effective when the labeled data is scarce compared with the unlabeled data. In active learning, the number of samples to learn a concept (or hypothesis) can be much less than the number required in classic supervised learning. Normally, there are three query strategies in active learning. (1) Query by uncertainty sampling \cite{proceeding10}, in which a learner selects the sample that is the most uncertain to label. (2) Query by Committee \cite{proceeding11}, in which the query sample has the most disagreement among a committee of models. (3) Query by generalization error \cite{proceeding12}, in which the query sample has the largest generalization error. In our work, the reconstruction error and sparse representation based clarification error are considered as the query criteria of active learning.

The contributions of this paper are summarized as follows:
\begin{itemize}
\item A new dictionary learning method via active learning in sparse representation is presented. In particular, the construction errors and sparse representation based classification errors are used as query criteria in the active learning.
 \item The proposed dictionary is applied to UCI \cite{url1} data sets (binary-category and multiple-category) and face data set, the reconstruction results and classification results are obtained in the process of sparse coding.
 \item  The proposed dictionary learning method is compared with other methods, clustering based dictionary, dictionary learning with structure incoherence, whole-training-data dictionary, and so on. The experimental results on the UCI data sets and face data set demonstrate the capability and efficiency of the proposed dictionary learning method.
\end{itemize}

The rest of the paper is organized as follows: Section 2 presents related work on dictionary learning, sparse representation based classification and query strategies in active learning. Section 3 presents the proposed dictionary learning via active learning. Section 4 presents the experiments with UCI data sets. Section 5 presents the comparison results on the Extended Yale B data set.  Section 6 gives the conclusion of the paper and discusses some future plans.

\section{Related Work}

 Assume a training data set T ($y_{1},y_{2},\cdots,y_{i},\cdots, y_{i}\in{R}^{m}$), the labels for the training data are L ($l_{1},l_{2},\cdots,l_{i},\cdots $).  In the original form of sparse representation, the sparsity of $x$ is defined as the number of nonzero elements. The constrain is named $\ell_0$ norm :

\begin{equation}
\label{eq:basic_L0}
min \|x\|_0, s.t. \hspace{3mm} Dx = y
\end{equation}

However, solving $x$ is NP hard. Fortunately, under Restricted Isometry Property (RIP) constrains, the solution of $\ell_0$ norm would be equivalent to $\ell_1$ norm. Then the sparse representation can be expressed as:
\begin{equation}
\label{eq:basic_L1}
min \|x\|_1, s.t. \hspace{3mm} Dx = y
\end{equation}
Under the $\ell_1$ norm, the solution becomes a convex optimization problem and $\ell_{1}$-regularized least squares method \cite{jour41} is used to solve this problem.
 \begin{equation}
\widehat{x}=arg\min\{\| y-Dx\|^{2}_{2}+\lambda\|x\|_{1}\}
\end{equation}

The choice of the dictionary $D$ is a key for effective sparse representation. There are plenty of researches on this topic in the literature. In the analytic approach, some pre-defined functions are used to construct the dictionary. Curvelets \cite{jour10}, which tracked the shape of the discontinuity set, supplied efficient and near-optimal representation of smooth objects. Shearlets \cite{proceeding13}, which obtained from dilations, action of translations, and shear transformations, displayed the geometric properties and mathematical properties for image representation. Bandelets \cite{jour12}, which specify the geometry as a vector field, improved image compression and noise reduction performance.

In the learning-based approach, machine learning methods are used to construct the dictionary from the training data. The least square error was used by the method of optimal directions (MOD) \cite{jour5} to update the dictionary iteratively:
   \begin{equation}
D_{k}=\arg\min_{D}\parallel Y-D_{k}X_{k}\parallel^{2}
\end{equation}
where k is the $k$-th iteration, Y is the training data matrix and $X_{k}$ is the sparse vector matrix based on the $k$-th dictionary. In KSVD \cite{jour2}, the atoms in the dictionary were updated sequentially. It was related to the k-means method and the atoms were modified based on associated examples. Online dictionary learning \cite{proceeding2}, which was based on stochastic approximations, adapted the dictionary to large data sets with millions of samples. In efficient sparse coding algorithms \cite{proceeding7}, two least square optimization problems ($\ell_1$ norm regularized and $\ell_2$ norm constrained) were solved interactively.
\begin{equation}
\label{eq:efficientcoding}
\min_{X, D}\frac{1}{2\sigma^2}\|DX-Y\|^2+ \lambda \|\alpha\|_1
\quad subject\hspace{2mm}to\hspace{2mm}\Sigma D\leq c
\end{equation}
where $\sigma, \lambda, \alpha$ and c are defined parameters. In the learning process, this approach optimized the dictionary $D$ or the sparse vector matrix X while the other is fixed.

In sparse coding, reconstruction error is the difference between the original testing data $y$ and the result of sparse representation, which can be expressed as:
   \begin{equation}
error_{y}=\sqrt[2]{\| y-Dx\|^{2}}
\end{equation}
and it is an important criteria to evaluate the quality of the dictionary.
\begin{algorithm}
 \caption{Sparse Representation based Classification}
 \begin{algorithmic}[1]
 \STATE {\bfseries Input:}  a test data $y\in \mathbb{R}^{m}$, a dictionary $D \in \mathbb{R}^{m\times n}$ with $c$ categories data
 \STATE Solve $\ell_{1}$-regularized least squares equation:\\
$\widehat{x}=arg\min\{\| y-Dx\|^{2}_{2}+\lambda\|x\|_{1}\}$
\STATE Calculate the residuals based on categories:

 $r_{i}(y)=\|y-D\delta_{i}(x)\|_{2}\quad for\quad i=1,\cdots,c$

 \STATE {\bfseries Output:} $label(y)=arg \min r_{i}(y)$
\end{algorithmic}
\end{algorithm}

Recently, sparse representation based classification (SRC) \cite{jour4} was proposed and presented with successful application in face image classifications. In SRC, the reconstruction errors based on different categories are used to classify testing data. For each class $i$, a function is defined as $\delta_{i}$, which selects the sparse vectors (coefficients) associated to $i$-th category. Then the SRC process can be presented as:
 \begin{equation}
label(y)=arg \min r_{i}(y),\quad r_{i}(y)=\|y-D\delta_{i}(x)\|_{2}
\end{equation}
The classification accuracy is utilized as another criterion to evaluate the property of the dictionary. The process of SRC is shown in Algorithm 1.

All active learning methods involve assessing the information of training data. The most informative sample $y^{*}$ is chose according to diverse query strategies \cite{jour42}. Query by uncertainty sampling \cite{proceeding10} and query by generalization error \cite{proceeding12} are two classical strategies for active learning.

In the uncertainty sampling, an active learner selects the sample which is least confident of labeling. And this method is usually straightforward in probabilistic learning models. Generally, the uncertainty sample is decided by the posterior probability:
 \begin{equation}
y^{*}=arg \max 1-P_{\theta}(\hat{l}|y)
\end{equation}
where $\hat{l}=arg \max_{l}P_{\theta}(\hat{l}|y)$, the classification is decided by the highest posterior probability with the model $\theta$. However, the least confident method simply evaluates the most uncertain data, which ignores the information of the rest data. The margin sampling method is proposed in \cite{proceeding14} with multi-class uncertainty criterion.
 \begin{equation}
y^{*}=arg \min P_{\theta}(\hat{l_{1}}|y)-P_{\theta}(\hat{l_{2}}|y)
\end{equation}
where $\hat{l_{1}}$ and $\hat{l_{2}}$ are the first and second most probable classification labels with the model $\theta$. Margin sampling method aims to utilize the posterior probability of the second most likely label. The sample with small margins are difficult for classifier to make decision. Therefore, obtaining the true label would improve the discriminated capability of model effectively. The most popular uncertainty sampling strategy uses entropy to evaluate training data:
 \begin{equation}
y^{*}=arg \max -\sum_{i}P_{\theta}(\hat{l_{i}}|y)\log P_{\theta}(\hat{l_{i}}|y)
\end{equation}
where $\hat{l_{i}}$ denotes possible labels. The entropy-based queries are successfully applied in complicated structured samples, such as trees \cite{proceeding15} and sequences \cite{proceeding16}.

In the case of generalization error, the key idea is to estimate the future error based on the new training data $\mathcal{L}\cup <y, l>$. The samples with the minimal expected error would be selected in the active learning. One typical method is to minimize the 1/0-loss:

 \begin{equation}
y^{*}=arg \min_{y}\sum_{i}P_{\theta}(l_{i}|y)(\sum_{u=1}^{U}1-P_{\theta^{+<y,l_{i}>}}(\hat{l}|y^{(u)}))
\end{equation}
where $\theta^{+<y,l_{i}>}$ stands for the new model with the new data $<y, l>$ added to the $\mathcal{L}$.

\section{Active Dictionary Learning (ADL)}
In active learning scenario, the query data $y^{*}$ is usually the most informative data from the unlabeled data pool. In the dictionary learning of sparse representation, if the atoms in the dictionary are the most informative samples in the training data, the dictionary would be representative and meaningful in the coding process.

\begin{algorithm}
 \caption{Active Dictionary Learning (ADL)}
 \begin{algorithmic}[1]
 \STATE {\bfseries Input:} A training data set T ($y_{1},y_{2},\cdots,y_{i},\cdots,y_{N}, i=1,2,,\cdots,N, y_{i}\in{R}^{m}$), the labels for the training data L ($l_{1},l_{2},\cdots,l_{i},\cdots,l_{N} $), the target number of the atoms $n$ in the dictionary\\

 \FOR{$k=1,2,\cdots,K$}
 \STATE Randomly choose $n$ data from T to establish a dictionary $D_{k} \in \mathbb{R}^{m\times n}$ \\
\STATE Calculate the sparse vector $x_{i}$ for each data $y_{i}$ based on dictionary $D_{k}$:\\
\begin{equation}
\widehat{x_{i}}=arg\min\{\| y_{i}-D_{k}x_{i}\|^{2}_{2}+\lambda\|x_{i}\|_{1}\}
\end{equation}
\STATE Calculate the reconstruction error for each data $y_{i}$:\\
$error^{r}_{(k,i)}=\sqrt[2]{\| y_{i}-D_{k}x_{i}\|^{2}}$
\STATE Calculate the classification error for each data $y_{i}$ based on Algorithm 1.
   \IF{$label(y_{i})=l(i)$} \STATE $error^{c}_{(k,i)}=0$
    \ELSE \STATE$error^{c}_{(k,i)}=\eta\times mean(error^{r}_{(k,i)})$
    \ENDIF
\ENDFOR
\STATE The total errors $E(y_{i})$ for each data $y_{i}$ after K test:\\
  \begin{equation}
  E(y_{i})=sum(error^{r}_{(k,i)})+sum(error^{c}_{(k,i)})
    \end{equation}
\STATE The data, which have the largest $E(y_{i})$, are selected for ADL:
  \begin{equation}
           y^{*}=\arg\max_{i}E(y_{i})
  \end{equation}

 \STATE {\bfseries Output:} Dictionary $D_{al}$ which have $n$ largest $E(y_{i})$ data
\end{algorithmic}
\end{algorithm}

In the proposed ADL method, the reconstruction $error^{r}$ and the sparse representation based classification $error^{c}$ are used to select the most the informative data in the training data:
   \begin{equation}
y^{*}=\arg\max_{y}error^{r}(y)+error^{c}(y)
\end{equation}
This idea shares the properties of query by uncertainty sampling \cite{proceeding10} and query by generalization error \cite{proceeding12} in the active learning paradigm.

Algorithm 2 shows the details of proposed method. At first, a random dictionary is established for testing. Then the reconstruction error and SRC error are calculated for each data. And the process is carried out with K times, which can reduced the selection bias. Finally, the total errors based on K iterations are recorded for selecting the atoms. For the classification error, the result is right or wrong, which is a binary output. In our method, we use the mean error of the reconstruction errors to normalize the classification error, the details are shown in step 10 of Algorithm 2. $\eta$ is an empirical index, which can be modified with different requirements. In our experiment, $K=5$ and $\eta=5$ are used for empirical study.

The proposed method is related to the AdaBoost.M1 \cite{jour51}. In AdaBoost.M1, the weights of the training data ($x_i, y_i, i=1,2,\cdots,N$) is initialized as $w_i=1/N$, then a sequence of weak classifiers $G_k(x), k=1,2,\cdots,K$ are applied on the training data using corresponding weights $w_i$. At each step, the training data that were misclassified at the previous step have their weights increased, whereas the weights are decreased when the training data were classified correctly. And the update process is via:
   \begin{equation}
w_i\leftarrow w_i\times\exp[\alpha_k\times I(y_i\neq G_k(x_i))],i=1,2,\cdots,N
\end{equation}
where $\alpha_k$ is the normalized parameter at each step. In our method, the reconstruction errors and the classification error based on the different random dictionary $D_k$ are used to rank the training data, which is similar to the weights updates in the AdaBoost.M1.

\section{Experiments on UCI Data Sets}
The experiments on different UCI \cite{url1} data sets are shown in this section. Both binary-category data sets and multi-category data sets are used in the experiments. The proposed dictionary are more effective in the multi-category data sets, which is normally difficult in applications. The performances on the reconstruction and classification are shown for each data sets.

\subsection{UCI Data sets}
Nine UCI data sets are used in the experiments. The detail information are shown in Table \ref{table1}, which are the size of data, the category properties and the feature number of data. Data ``car evaluation" and ``vowel recognition" are binary-category data sets. Data ``contraceptive method choice", ``wine" and ``cardiotocography" have three categories. The rest are classic multi-category data sets.
\begin{table}
\renewcommand{\arraystretch}{1.3}
\caption{UCI Data Sets in Experiments}
\label{table1}
\centering
\resizebox{8.5cm}{!} {
\begin{tabular}{c|c|c|c }
\hline
Name & Feature number & Total size & Category number  \\
\hline	
car evaluation & 6 & 1728& 2\\	
vowel recognition& 10 &528&2\\
contraceptive method choice &9 & 1473& 3\\
wine &13& 178& 3\\
cardiotocography &22& 2126&	3\\
glass  &   10   &   214 &  7  \\
image segmentation& 19 & 2310 &7\\
libras movement   &   90   &   360 &   15  \\
breast tissue &9 &106&6\\
\hline
\end{tabular} }
\end{table}

\subsection{Experiment Setting}
For each data set, 5-fold cross-validation method is used. 80\% data are used as training data to establish dictionary, the rest 20\% data are left for testing. In order to investigate the performance of different sizes of dictionary. The sizes of dictionary are from 10\% (0.1) to 50\% (0.5) of the size of training data. Then according to Algorithm 2, the dictionary based on the active learning is trained.

In order to show the effectiveness of the proposed method, relative methods are utilized for comparisons. Two clustering based dictionary learning methods, self organized map (SOM) based dictionary and neural gas (NGAS) based dictionary, are used in the comparisons. SOM and NGAS are classic unsupervised learning methods, which can maintain the topological properties of the training data. Some sparse coding applications with NGAS are discussed in \cite{jour6}. The centers trained by SOM and NGAS are used as atoms in our experiments. The labels of centers are based on the 5-nearest neighbor voting. Whole-training-data dictionary, is used as standard in the experiments. For the classification, the SRC method is used among different dictionary learning models. Our sparse coding tool is from \cite{jour41}. The SOM and NGAS are from the SOM Toolbox \cite{ur3}.

For simplicity, ADL, SOMD, and NGASD are used to represent active dictionary learning, SOM based dictionary, and NGAS based dictionary separately. Whole-training-data dictionary is abbreviated as WD. It is important to note that WD contains all the training data in its dictionary.

\subsection{Result and Discussion}

\begin{figure*}
 \centering
\subfigure{\includegraphics[width=0.4\linewidth,height=2in]{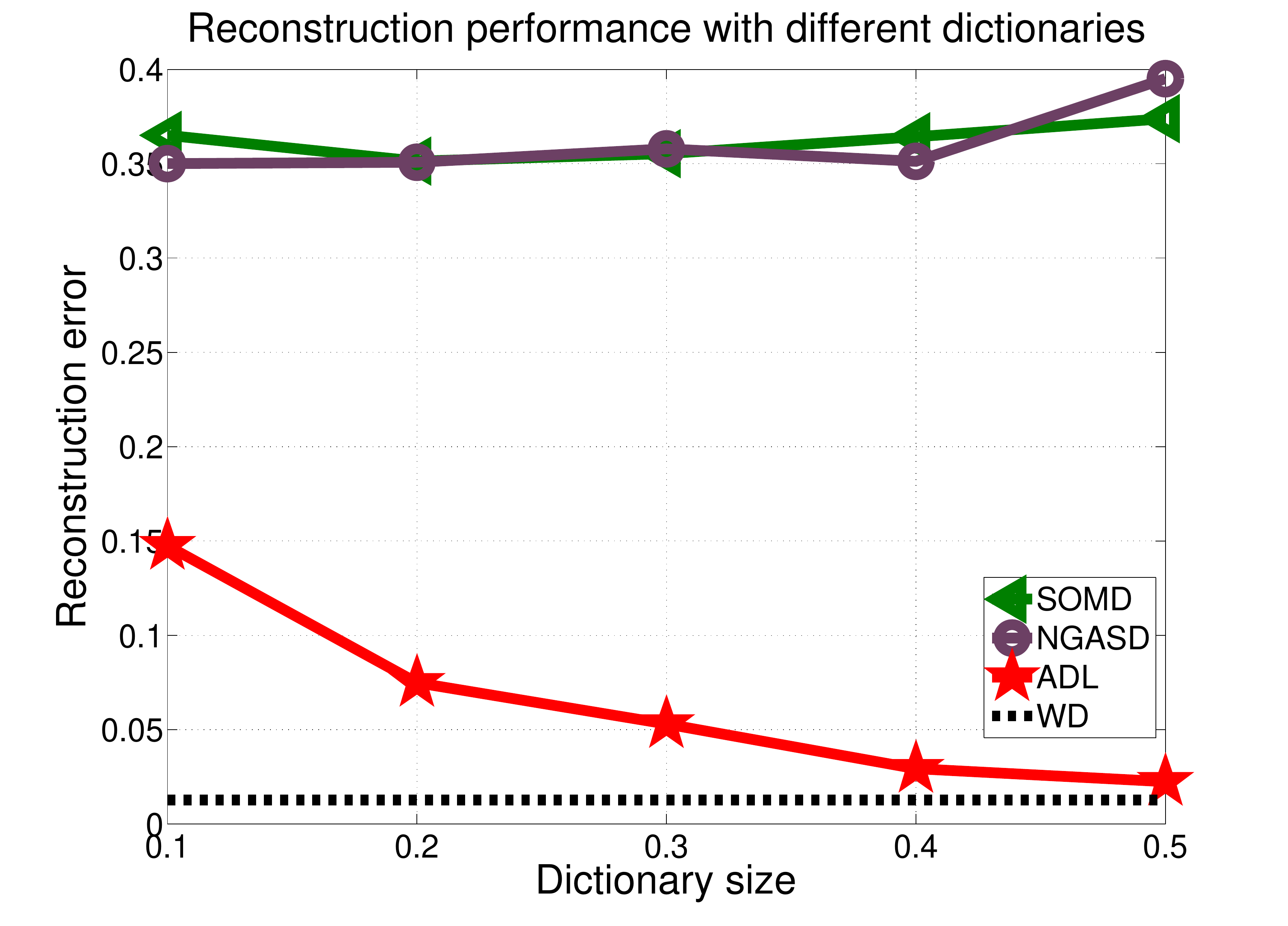}}
\hfil
\subfigure{\includegraphics[width=0.4\linewidth,height=2in]{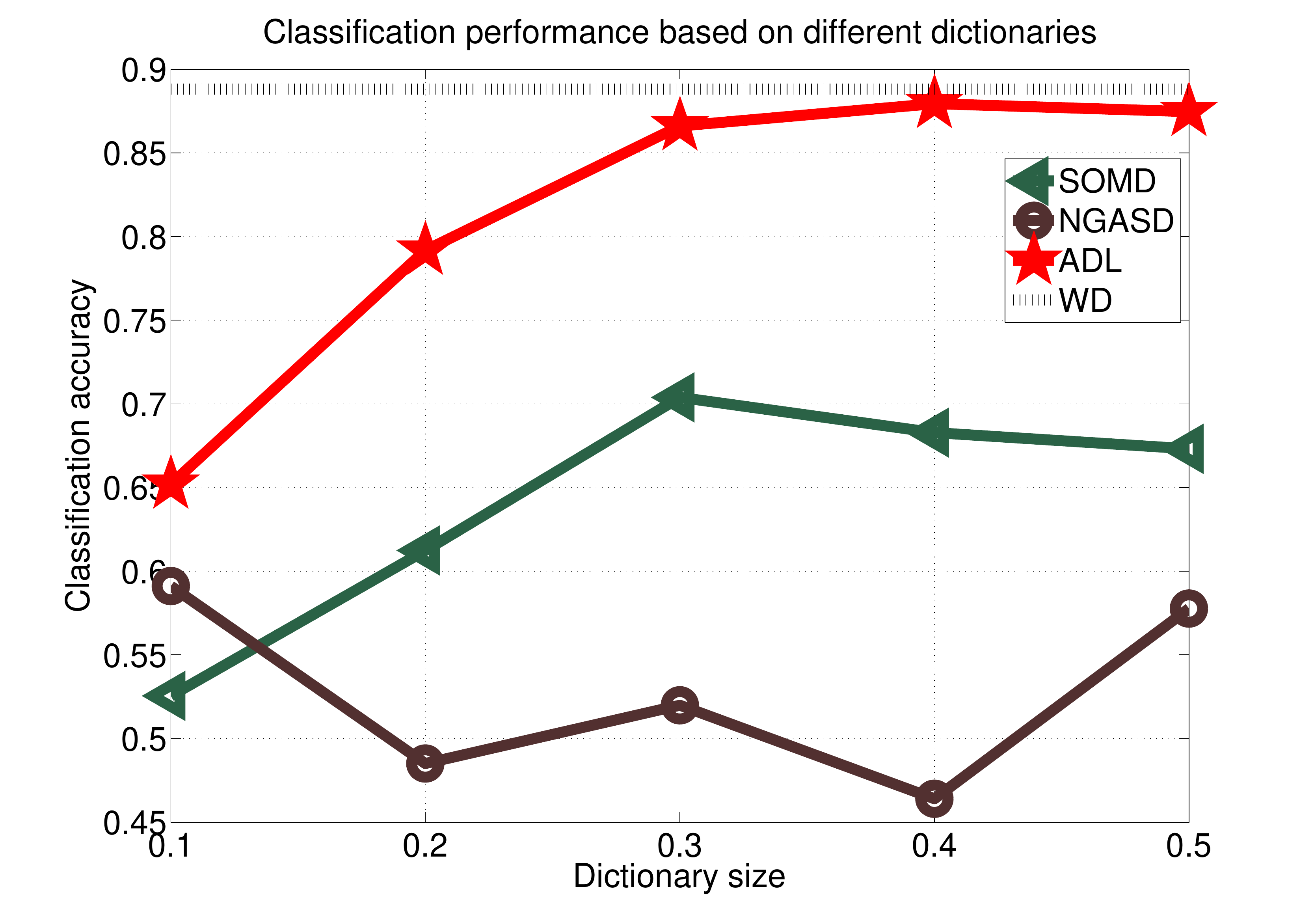}}
\hfil
\caption{$\mathbf{Left:}$ Reconstruction performance with data ``car evaluation" $\mathbf{Right:}$ Classification performance with data ``car evaluation"  }
\end{figure*}

\begin{figure*}
 \centering
\subfigure{\includegraphics[width=0.4\linewidth,height=2in]{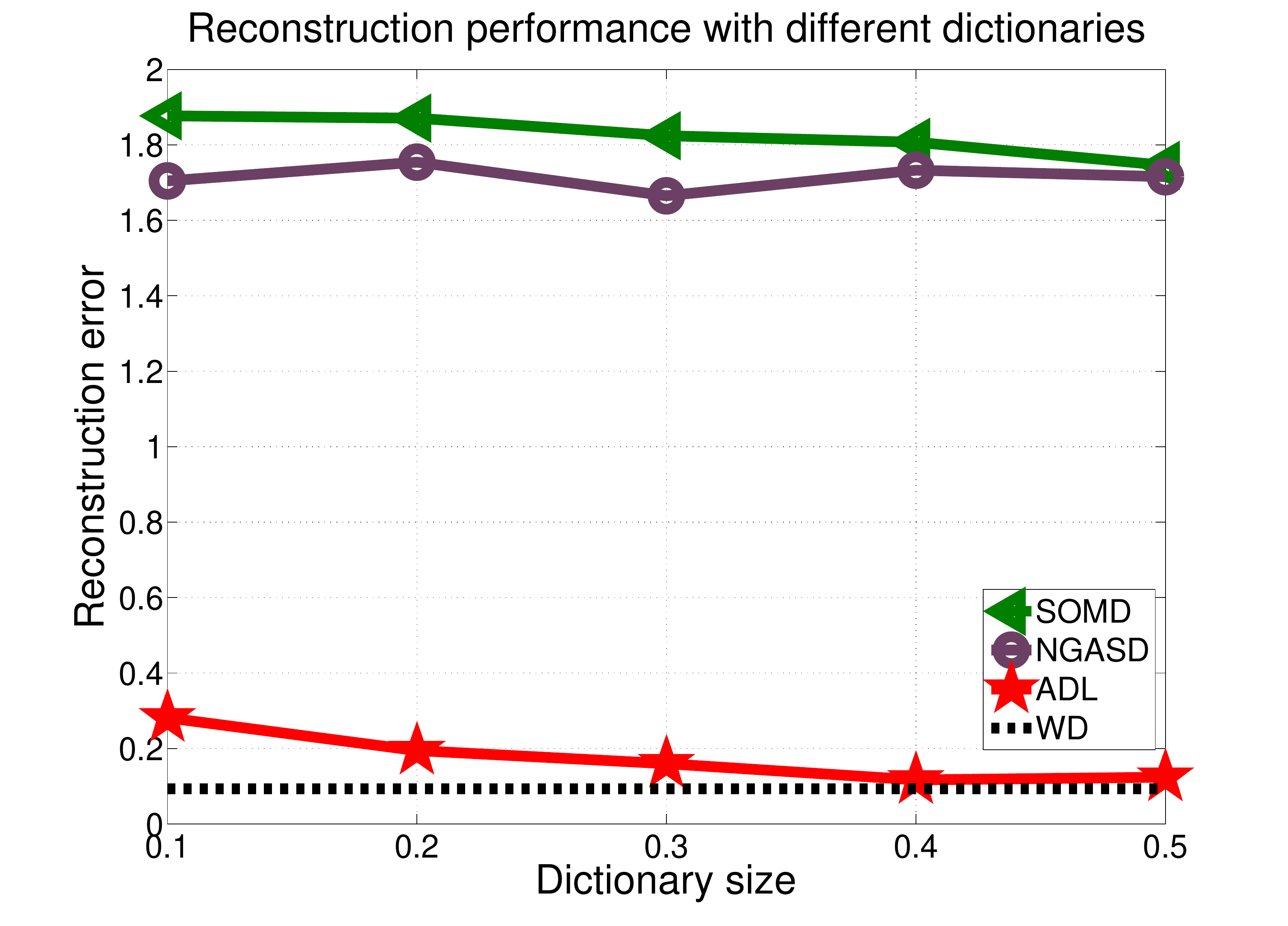}}
\hfil
\subfigure{\includegraphics[width=0.4\linewidth,height=2in]{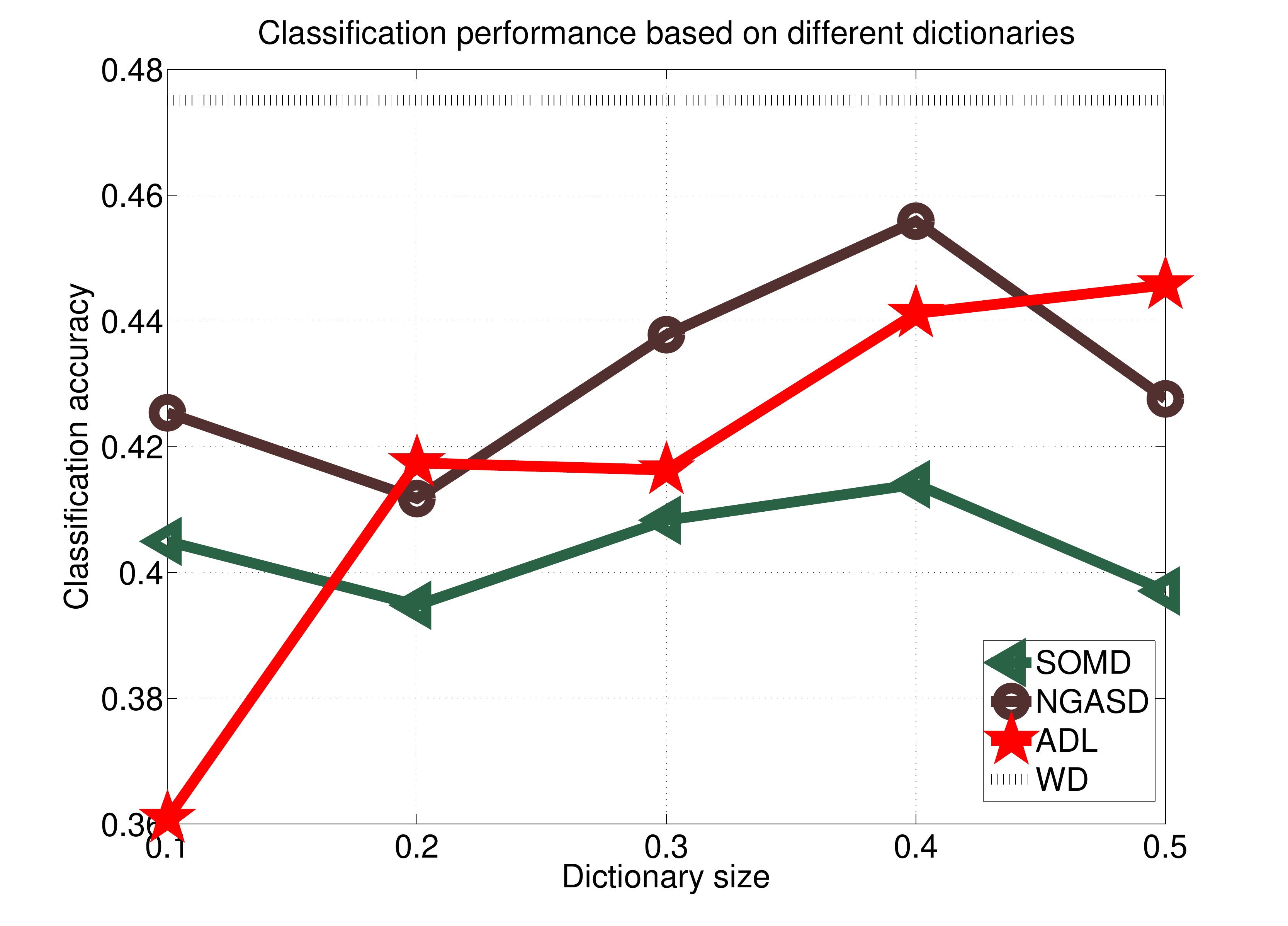}}
\hfil
\caption{$\mathbf{Left:}$ Reconstruction performance with data``Contraceptive Method Choice" $\mathbf{Right:}$ Classification performance with data ``Contraceptive Method Choice"  }
\end{figure*}

\begin{figure*}
 \centering
\subfigure{\includegraphics[width=0.4\linewidth,height=2in]{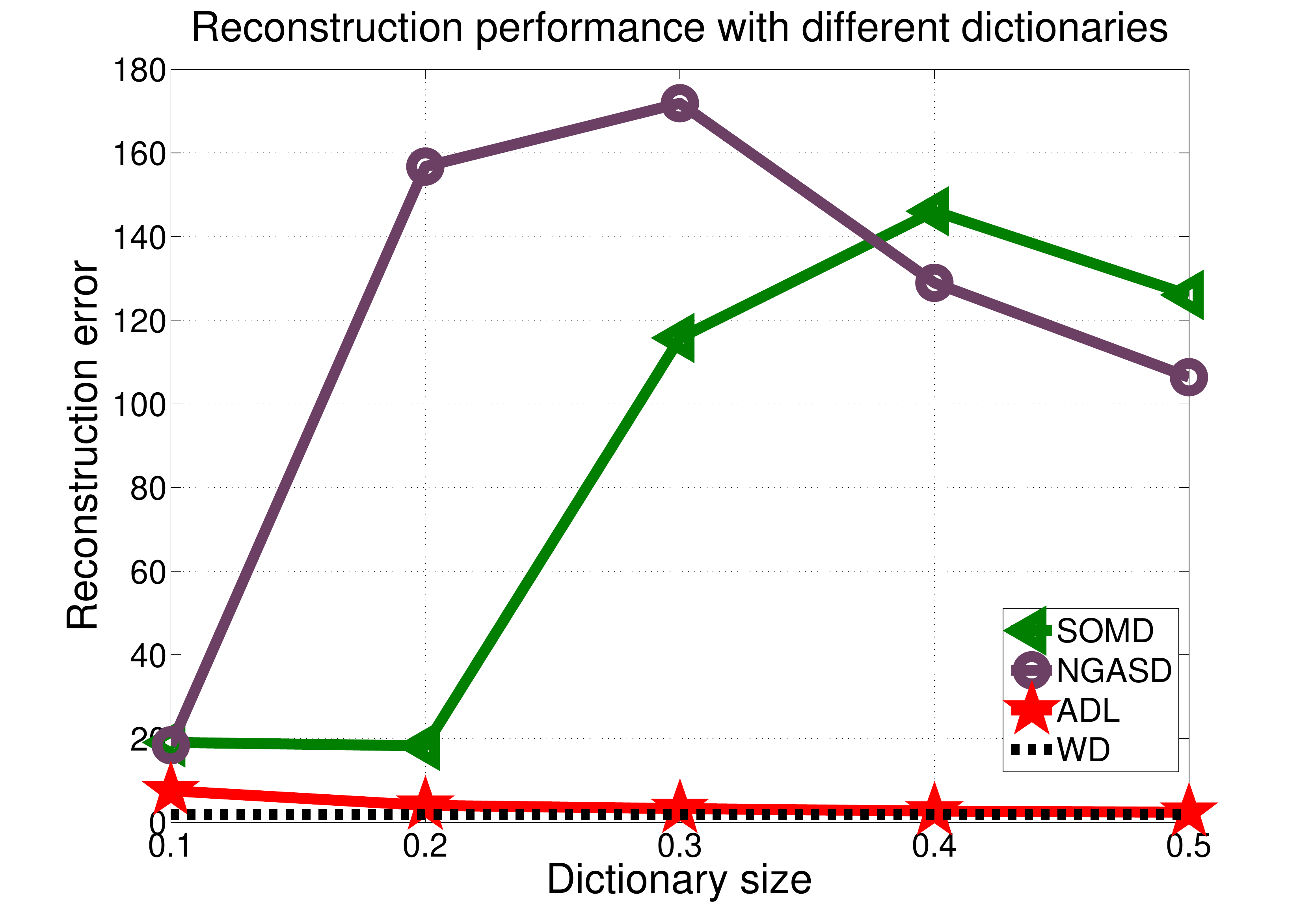}}
\hfil
\subfigure{\includegraphics[width=0.4\linewidth,height=2in]{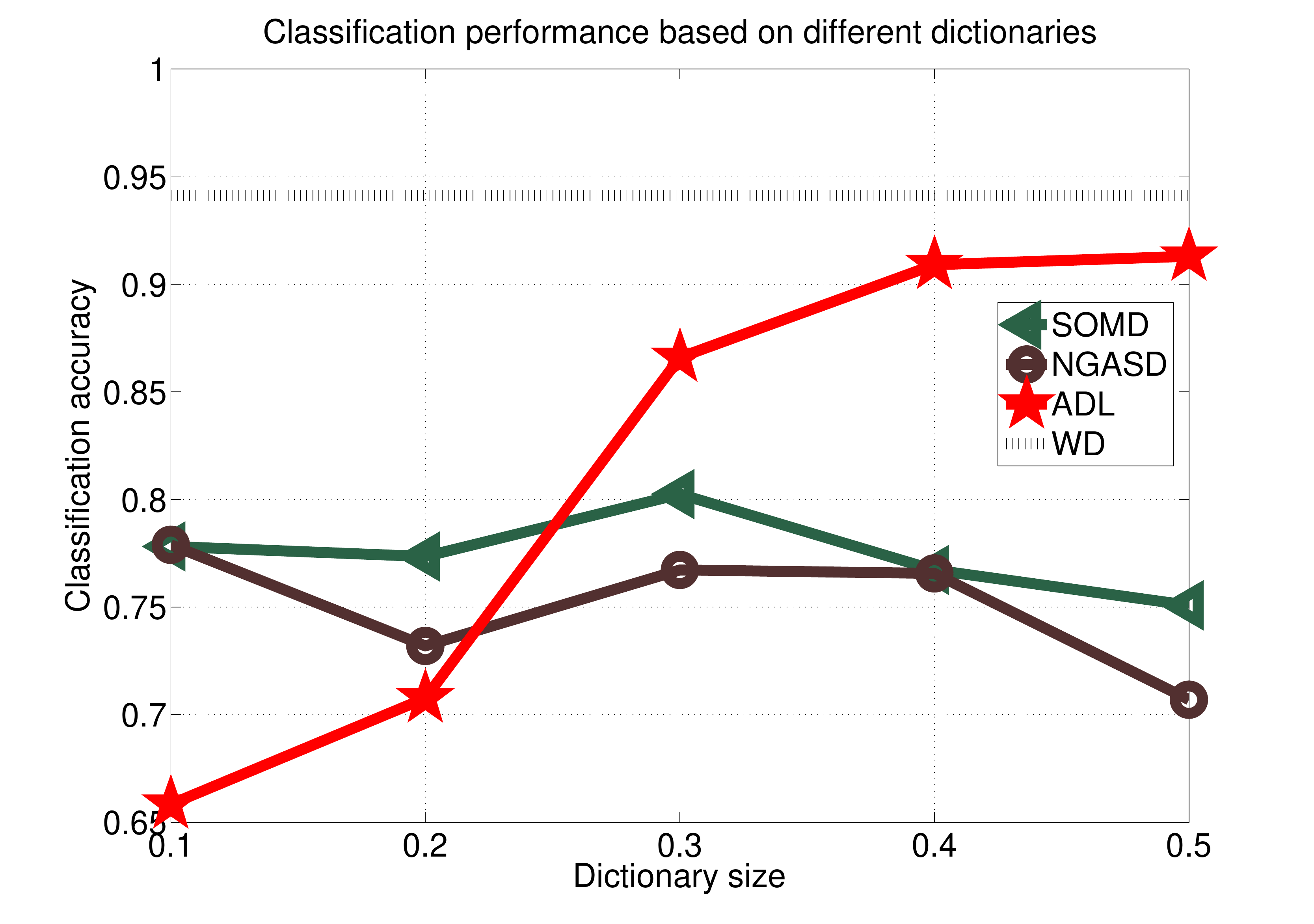}}
\hfil
\caption{$\mathbf{Left:}$ Reconstruction performance with data ``cardiotocography" $\mathbf{Right:}$ Classification performance with data ``cardiotocography"  }
\end{figure*}

\begin{figure*}
 \centering
\subfigure{\includegraphics[width=0.4\linewidth,height=2in]{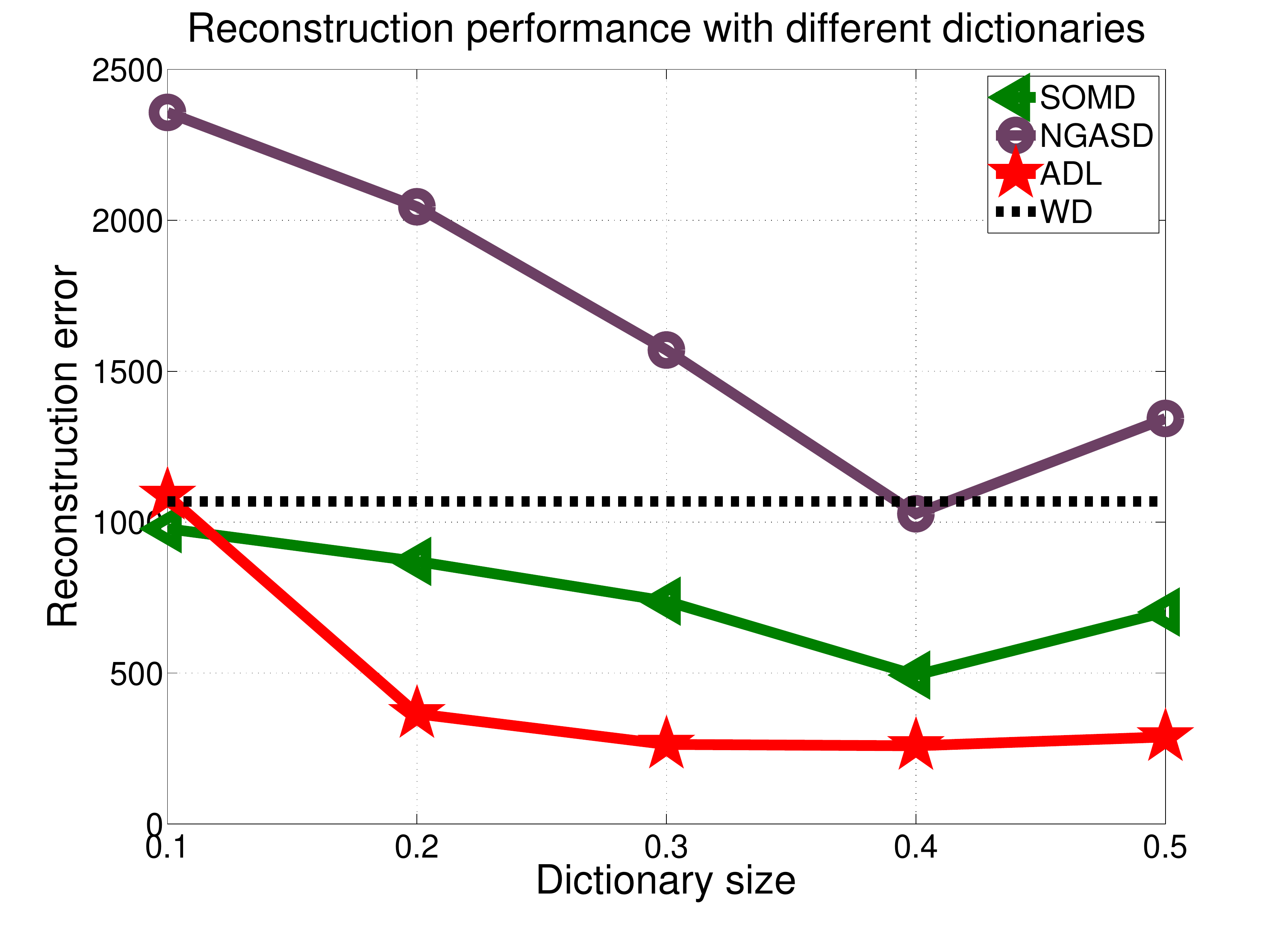}}
\hfil
\subfigure{\includegraphics[width=0.4\linewidth,height=2in]{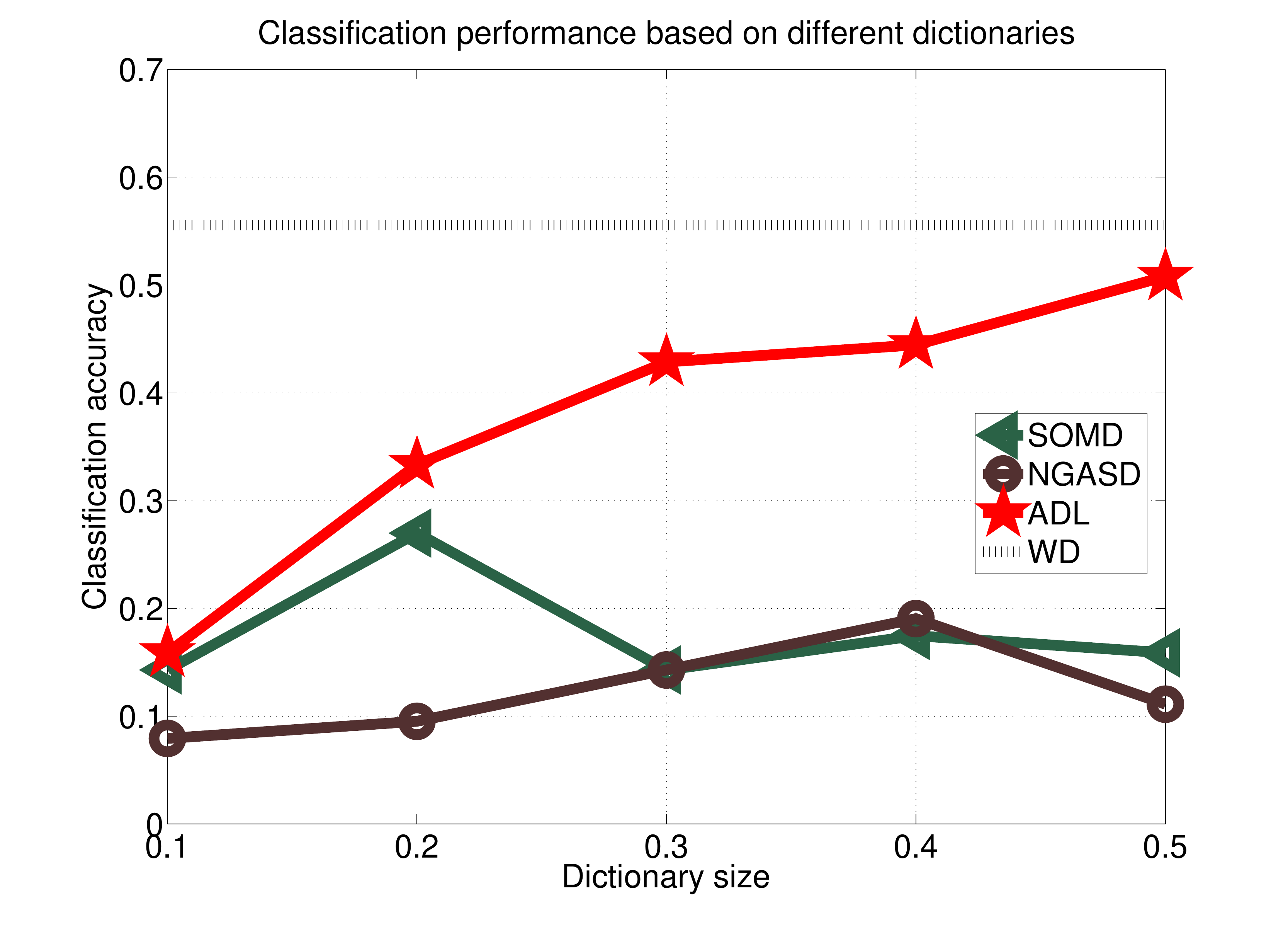}}
\hfil
\caption{$\mathbf{Left:}$ Reconstruction performance with data ``breast tissue" $\mathbf{Right:}$ Classification performance with data ``breast tissue"}
\end{figure*}

In this section, the comparison results based on reconstruction and classification are shown. We list the detail performances of 4 data sets. ``Car evaluation" is a binary-category data set, ``Contraceptive Method Choice" and ``cardiotocography" are three-category data sets, ``breast tissue" data set has more than three categories. The performances are based on different dictionaries size rate, i.e. 0.1, 0.2, 0.3, 0.4 and 0.5, which indicates the number of atoms in dictionary compared with the total number of training data. The reconstruction errors are the average errors on the testing data, which have different scales due the different data sets properties. Then the average results based on all 9 data sets are shown for comprehensive performance comparison.

Figure 1 shows the performances on the data ``car evaluation". For the reconstruction results on the left subfigure, ADL has relative smaller reconstruction errors. When the dictionary size rate are more than 0.4, the reconstruction errors of ADL are comparable to the level of WD. For the classification performance, ADL always has the highest accuracy. When the dictionary rates are larger than 0.3 the results of ADL reach the level of WD results.

Figure 2 shows the performances of the three-category data ``Contraceptive Method Choice". For the reconstruction results, the curve of ADL is at the bottom of all the dictionary learning methods, and the errors are comparable with WD when the dictionary size rates are 0.4 and 0.5. For the classification results on the right subfigure, ADL has competitive performance with NGASD from the rate 0.2.

The results of another three-category data ``cardiotocography" are shown in Figure 3. In the left subfigure, ADL has small reconstruction errors comparable to the WD. For the classification results, ADL has higher accuracies when the dictionary size rates are 0.3, 0.4 and 0.5.

Figure 4 shows the performance of data ``breast tissue". For the reconstruction results, the error curve of ADL is at the bottom of all the models from the dictionary size rate 0.2. It is interesting to note that the error of WD is high. For the classification performance, the results of ADL rank first compared with SOMD and NGASD.

In the above figures, ADL has shown some advantages in the reconstruction and classification compared with relative methods. Table \ref{table2} lists the average performances based on the all 9 data sets in our experiments. In detail, the mean accuracy with different dictionary size are shown among 3 dictionary learning models. The results show the accuracies of ADL rank first when the dictionary rate is larger than 0.1. For the reconstruction, as different data sets have different scales, it is not meaningful to take the average reconstruction error. For the mean ranks among 3 dictionary learning models,  we observer that ADL always ranks first with different dictionary size rates.

 \begin{table}
\renewcommand{\arraystretch}{1.4}
\caption{Mean accuracy rate [\%]in different dictionary sizes}
\label{table2}
\centering
 {
\begin{tabular}{c||c|c|c|c|c }
\hline
Dictionary size (rate) & 0.1&0.2 &0.3 &0.4 &0.5\\
\hline\hline	

SOMD &44.14 &50.18 &53.19 	&56.12 	&50.90\\
NGASD  & 46.95& 	46.47 &	46.24& 	46.28& 	44.51\\
ADL& 46.12& 	58.28& 	67.92 &	70.62 &	74.19\\
\hline
\end{tabular} }
\end{table}

\section{Experiments on Face recognition}
\begin{figure}
\centering
\includegraphics[width=3.4in]{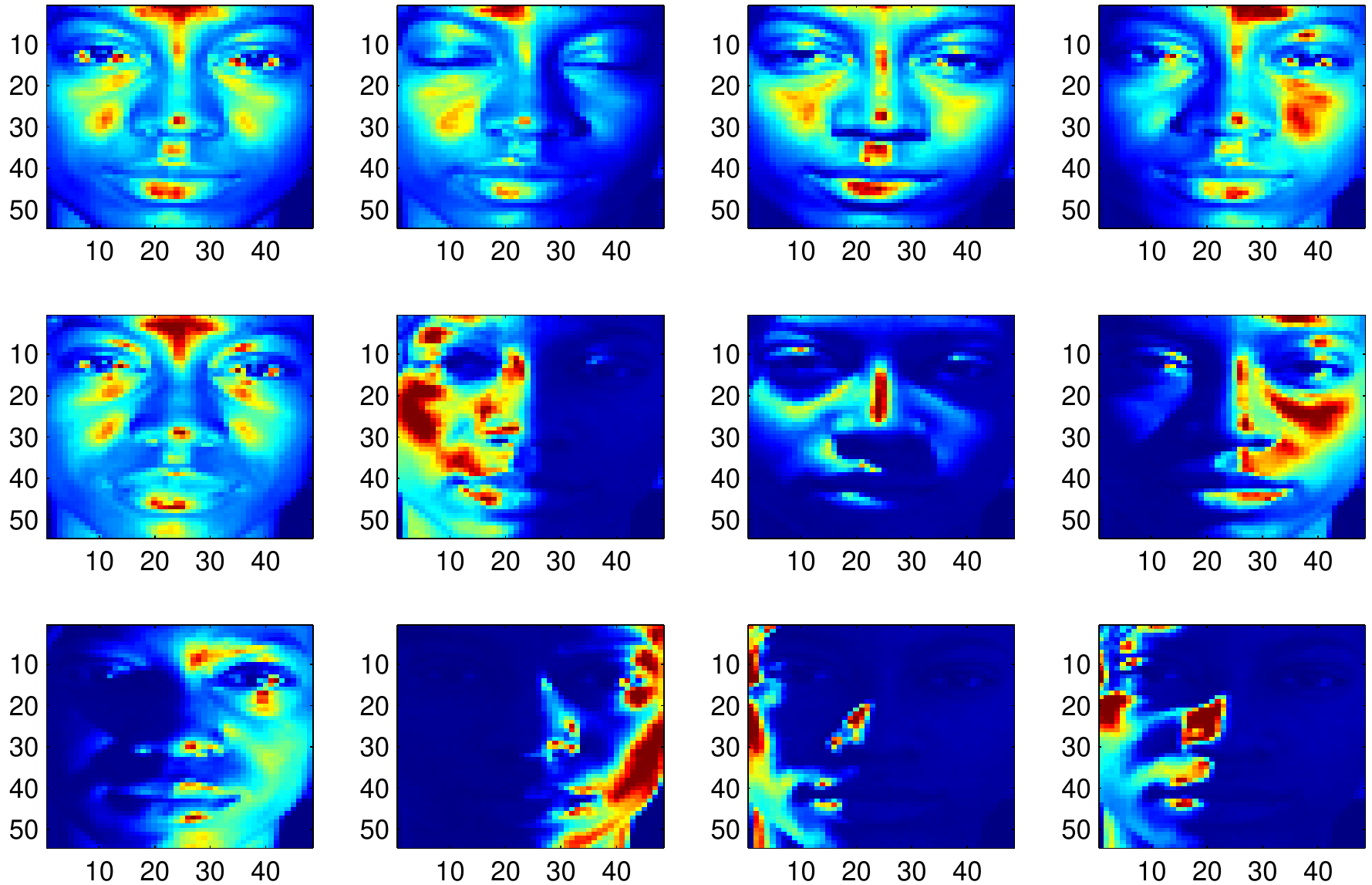}
\caption{The atoms source for a person in ADL dictionary (dictionary size is 600). The display uses the jet colormap from Matlab.}
\label{figface}
\end{figure}

 \begin{table}
\renewcommand{\arraystretch}{1.2}
\caption{The classification results on the Extended Yale B database mentioned in FDDL work}
\label{fddl}
\centering
 {
\begin{tabular}{c||c|c|c|c|c|c|c }
\hline
Methods&NN&DKSVD&DLSI&SVM&DLSI*&SRC&FDDL\\
\hline\hline	
Accuracy&0.62&0.75&0.85&0.89&0.89&0.90&0.92\\
\hline
\end{tabular} }
\end{table}

 \begin{table}
\renewcommand{\arraystretch}{1.4}
\caption{The classification results of ADL on the Extended Yale B database }
\label{adl}
\centering
 {
\begin{tabular}{c||c|c|c|c }
\hline
Dictionary Size&600&650&700&750\\
\hline\hline	
ADL Accuracy&0.86&0.87&0.87&0.90\\
\hline
\end{tabular} }
\end{table}

Recently, Fisher Discrimination Dictionary Learning (FDDL) was proposed \cite{proceeding17} and successfully applied in the extended Yale B Face data set \cite{jour7}. In this data set, there are 2414 image faces from 38 persons (64 images for each person). In the setting of the FDDL, the dictionary selected 20 images from each person. There are $38\times20=760$ images in the training data and the rest are testing data. There are 6 competitive methods showing in \cite{proceeding17} for comparison and results are shown in Table \ref{fddl}. For details, SRC (sparse representation based classification), two classical classifiers: NN (nearest neighbor) and SVM (linear support vector machines), two new proposed dictionary learning based classification methods: DKSVD (discriminative KSVD) \cite{proceeding18} and DLSI (dictionary learning with structure incoherence) \cite{proceeding19}. Note that there are two DLSI (DLSI and DLSI*) used in that work. The atoms number of dictionary are 760 in the methods related to the sparse representation (such as SRC, DKSVD and DLSI).

When we use ADL on the same data set, we keep the experiment setting as FDDL (Eigen face with dimension 300) and the outputs are the average of 5 times running. However, we don't select the training data from each class (person), instead, we randomly choose 760 images as the training data from the data set. The reason lies in two aspects: First, choosing training data with each class is cost of labor, especially in the huge date set. Second, we try to test the robustness of our method in the imbalance learning framework \cite{jour8}.

 Figure~\ref{figface} shows the atoms source for a person when the ADL dictionary has 600 atoms. In the ADL dictionary, the size of an atom is 300 after Eigen face process and we have traced back to display the original face image for each atom. The selected face images are representative of different face expressions and light conditions. We show our classification results with dictionary size of 600, 650, 700 and 750 in Table \ref{adl}. The accuracies we got are higher than that of NN, DKSVD and DLSI, and they are in the same level with SVM, DLSI* and SRC. The results of ADL is slight lower than that of FDDL. However, our method can use only one dictionary to handle all the classes while there are many dictionaries based on diverse classes in FDDL.

\section{Conclusion}
A novel dictionary learning method based on active learning is proposed in the paper. It is an iterative searching criteria in the training data. Compared with classic active learning, which is active to search for data to label, ADL is active to search for data to establish dictionary. The comparisons with other dictionary learning methods give the comprehensive study. The results show that ADL is effective in reconstruction and classification due to the small dictionary size and randomly sampled training data. In certain cases, ADL with small size dictionary can achieve comparable performance with whole-training-data dictionary.

Theoretical analysis and more data sets are needed to extend the proposed dictionary learning method. How to balance the rate between the reconstruction error and classification error is still an open question in the sparse representation. Our method has shown to be effective for this problem in the dictionary learning. Beyond this, we plan to modify our method for large data set applications, such as social networks analysis \cite{jinadd7}, \cite{jinadd8} and human group recognition \cite{jinadd9}.




%

\end{document}